\pdfoutput=1

\documentclass[11pt]{article}

\usepackage[final]{acl}

\usepackage{times}
\usepackage{latexsym}

\usepackage{soul}

\usepackage[T1]{fontenc}

\usepackage[utf8]{inputenc}

\usepackage{microtype}

\usepackage{inconsolata}

\usepackage{graphicx}

\usepackage[colorinlistoftodos]{todonotes}

\title{A Computational Framework to Identify Self-Aspects in Text}

\author{
  Jaya Caporusso\textsuperscript{1,2} \quad
  Matthew Purver\textsuperscript{1,3} \quad
  Senja Pollak\textsuperscript{1} \\
  \textsuperscript{1}Jožef Stefan Institute, Ljubljana, Slovenia \\
  \textsuperscript{2}Jožef Stefan International Postgraduate School, Ljubljana, Slovenia \\
  \textsuperscript{3}Queen Mary University of London, United Kingdom \\
  \texttt{jaya.caporusso@ijs.si}
}

\begin{document}
\maketitle
\begin{abstract}

This Ph.D. proposal introduces a plan to develop a computational framework to identify Self-aspects in text. The Self is a multifaceted construct and it is reflected in language. While it is described across disciplines like cognitive science and phenomenology, it remains underexplored in natural language processing (NLP). Many of the aspects of the Self align with psychological and other well-researched phenomena (e.g., those related to mental health), highlighting the need for systematic NLP-based analysis. In line with this, we plan to introduce an ontology of Self-aspects and a gold-standard annotated dataset. Using this foundation, we will develop and evaluate conventional discriminative models, generative large language models, and embedding-based retrieval approaches against four main criteria: interpretability, ground-truth adherence, accuracy, and computational efficiency. Top-performing models will be applied in case studies in mental health and empirical phenomenology.

\end{abstract}

\section{Introduction}

The Self, superficially experienced as ``the (perhaps sometimes elusive) feeling of being the particular person one is'' \cite{siderits2013self}, is a complex phenomenon, amply discussed in philosophy and cognitive science \citep[e.g.,][]{zahavi2008subjectivity}.
While there exist different views about the metaphysical nature of the Self \cite{siderits2013self}, in this work, we build on its phenomenological and behavioural manifestations. In everyday experience, the Self is characterised by multiple phenomenological and psychological aspects, including the experience of one’s own body \cite{bermudez2018bodily} and a sense of agency \cite{gallagher2000philosophical}, among others \cite{caporusso2022dissolution}.

These Self-aspects are conceptually and empirically related to other well-established constructs—such as personality traits or experiential modes. For example, their relevance to contexts such as mental health research is supported in related work, which highlights the central role of Self-related processes in well-being and psychopathology, as well as in empirical phenomenology \citep[i.e., the empirical investigation of experience;][]{aspers2009empirical}, where they are key to understanding altered states of consciousness (see Section \ref{Sec:RelWork}).

Importantly, the specific ways in which Self-aspects are experienced by a person in a given moment are reflected in the language they use \citep[e.g., see Section \ref{Sec:RelWork} and][]{pennebaker2003psychological}. The found correlations between textual features and Self-aspects can be further employed in downstream NLP tasks, for instance to detect psychological states \cite{caporusso-etal-2023-ijs, du2022linguistic, kolenik2024computational}. However, the connections between textual features and many Self-aspects important for the identification of, e.g., mental health conditions and phenomenological states, are underexplored.

To address this shortcoming, we propose a computational framework capable of automatically detecting the presence and mode of Self-aspects in text. Existing tools such as LIWC \citep[Linguistic Inquiry and Word Count;][]{boyd2022development} and VADER \citep[Valence Aware Dictionary and sEntiment Reasoner;][]{hutto2014vader}
have shown that psychologically meaningful patterns can be computationally extracted from text using lexicons and interpretable features. Building on this tradition, our framework aims to go further: to detect nuanced, theoretically grounded aspects of Self-experience—such as agency, embodiment, or narrative coherence—through a combination of ontology design, annotated data, and a range of modelling approaches. The resulting method can be applied to tasks in domains such as mental health research and empirical phenomenology.


\section{Related Work}
\label{Sec:RelWork}


\subsection{Textual Features and Self-Aspects Correlations}

This subsection surveys studies mapping text features to aspects of the Self.

\paragraph{Self-Aspects}
Most research focuses on \textit{I-talk}, i.e., the use of first-person pronouns as indicators of Self-focus \cite{pennebaker2003psychological}, which correlates with emotional pain, trauma, and depression \cite{tausczik2010psychological}.  Furthermore, pronoun usage hints at specific understandings of the Self vs others distinction \cite{na2009culture, sharpless1985identity}. The usage of active vs passive voice can shed light on the sense of agency of the author of a text \cite{simchon2023computational}, while the Narrative Self \citep[NS; i.e., ``the narrative someone has of themselves, comprising their autobiographical memories and stories of who they are''][]{caporussophenomenologically} is reflected in the structure and coherence of one's autobiographical accounts \cite{habermas2015autobiographical, holm2016life, jaeger2014trauma, waters2015relations}. In this context, Author profiling (AP) refers to the task of inferring personal characteristics of an author based on their writing, which has applications in, e.g., sociolinguistics and mental health analytics \cite{eke2019survey, ouni2023survey}. 

The correlation of text features with other aspects of the Self, such as the Minimal Self \citep[MS; ``the fact that experiences are presented to us in a fundamentally personal and subjective way''][]{caporussophenomenologically}, are less explored \cite{uno2025minimal}.  

\citet{caporussophenomenologically} investigated the LIWC categories associated with different aspects of the Self: MS, NS, Self as Agent (AS; ``the experience of being an agent, i.e., in control, active''), Bodily Self (BS; ``the experience of owning, controlling, and/or identifying with someone’s own body (or parts
of it)''), and Social Self (SS; ``the self as it is shaped and/or perceived when in an interaction or relationship of sorts with
other people or entities to whom we attribute qualities of an inner life''). Specifically, utilising a mixed approach to annotate the data, the authors classified text instances as presenting or not each of the mentioned Self-aspects, and they analysed the obtained splits with
LIWC.

\paragraph{Methods}
The methodological approaches utilised to detect correlations between textual features and Self-aspects can be broadly grouped into three main types:
\begin{itemize}
    \item Approaches based on stylistic features such as punctuation, syntactic patterns, part-of-speech (POS) tags, sentence length, character/word n-grams, and structural features (e.g., number of paragraphs or capitalised words)---see \citet{ouni2021toward, vijayan2019survey}.
    \item Content-based approaches, relying on subject matter and vocabulary; features include term frequency-inverse document frequency (TF-IDF), topic models, and domain-specific keywords---see \citet{ch2018study, ouni2023survey}.
    \item Hybrid approaches, where both stylistic and content-based features are analysed---see \citet{fatima2017multilingual, ouni2021toward, ouni2023survey}.
\end{itemize}

The use of LIWC or other lexicon-based techniques is the most common approach to investigate correlations between Self-aspects and textual features \cite{boyd2021natural, pennebaker2003psychological}. More recently, however, NLP research has increasingly adopted machine learning (ML) methods---such as topic modelling and supervised classification---to analyse language patterns in a data-driven way \cite{eichstaedt2018facebook, ouni2021toward}. Many studies used classical supervised learning methods, like support vector machines \citep[SVMs;][]{chinea2022zero, hacohen2022survey, vijayan2019survey}, random forests \citep[RFs;][]{fatima2017multilingual, ouni2021toward}, decision trees \cite{vijayan2019survey}, and Naïve Bayes \citep[NB;][]{mechti2020improving}. Feature extraction in AP is critical: common strategies include Bag-of-Words (BoW) and TF-IDF \cite{ouni2023survey}, character and word n-grams \cite{hacohen2022survey}, POS and syntactic feature vectors \cite{mechti2020improving, vijayan2019survey}, word embeddings \cite{chinea2022zero, fatima2017multilingual}, semantic graphs and emotion tags \cite{ouni2023survey}. Furthermore, many studies employ qualitative approaches \cite{habermas2015autobiographical, waters2015relations}. However, deep learning (DL) models are increasingly employed as well, due to their capacity to automatically learn hierarchical feature representations from raw text and their superior performance on large-scale NLP tasks \cite{ouni2023novel}. Transformer-based models such as BERT \cite{devlin2019bert} and RoBERTa \cite{liu2019roberta} were adapted to AP tasks by fine-tuning on labelled AP datasets \cite{chinea2022zero}. In recent work, large language models (LLMs) have been explored for AP \citep[see][]{huang2025authorship}. \citet{huang2024can} show that GPT-4 outperforms BERT-based models in zero-shot authorship attribution and verification, especially when guided by linguistic cues. 

The type of text analysed varies widely, ranging from autobiographical essays \cite{adler2012living, mcadams2001psychology}, stream-of-consciousness essays or narrative prompts \cite{pennebaker1986confronting, rude2004language}, transcripts of spoken conversations or interviews \cite{adler2008narrative, bamberg2008considering, lysaker2002narrative}, diary entries and letters \cite{baumeister1994guilt, pennebaker1996cognitive}, social media posts \cite{guntuku2019studying, schwartz2013personality}, to even published autobiographies or literature \cite{bruner2003making, freeman2009hindsight}.

\subsection{Downstream Applications}

The correlations discussed in the previous subsection are often employed in downstream applications. For instance, \citet{kolenik2024computational} utilised predefined sets of words and linguistic patterns that have been associated with specific psychological states, traits, or cognitive processes to train ML models that detect stress, anxiety, and depression. Similarly, \citet{du2022linguistic} leveraged linguistic features known to correlate with psychological states, like absolutist words and personal pronouns, to detect depression, anxiety, and suicidal ideation. In the context of the LT-EDI@RANLP 2023 shared task \cite{ltedi-2023-language}, first-person singular pronouns and time-related terms, recognised as indicative of depressive states \cite{ratcliffe2014experiences}, were employed to identify signs of depression in social media posts \cite{caporusso-etal-2023-ijs}. \citet{eichstaedt2018facebook} utilised topic models to identify clusters of words that often appear together in Self-narratives, and supervised ML to predict an upcoming depression diagnosis from social media posts.

Outside of the context of NLP studies, works investigating, e.g., mental health issues or phenomenological states, vastly address Self-aspects to identify the phenomenon of interest. For instance, an impacted sense of agency is registered in individuals with anxiety and depression, who experience a deficiency in estimating their control over positive outcomes \cite{mehta2023reward}, while disturbances in interoception and Self-awareness were found to be correlated with anxiety and schizophrenia, among the others \cite{yang2024interoception}
. Often, different Self-aspects correlate with disorders in a synergistic way, or there is an atypical disintegration of Self-aspects. For instance, Alzheimer's disease and other conditions involving cognitive decline are associated with impaired Self-continuity, sense of personal history and future goals, capabilities of Self-reflection, and personal meaning \cite{el2015autobiographical}, resulting in a distorted narrative Self-identity. Alongside---and sometimes in support of---research in mental well-being, Self-aspects are also relevant in the context of empirical phenomenology, among other domains. For example, a multitude of Self-aspects is examined in the investigation of experiences of dissolution \citep[i.e., "experiential episodes during which the perceived boundaries between self and world (i.e., nonself) become fainter or less clear"][]{caporusso2022dissolution, nave2021self}, and bodily experience is investigated in the context of depersonalisation and derealisation disorders \cite{tanaka2018like}. In line with this, scales and symptom checklists have been developed to assess the presence and intensity of psychological or phenomenological states \cite{heering2016disturbed, michal2014striking, nour2016ego, parnas2005ease, sierra2000cambridge}.

\subsection{Identified Gaps and Research Motivation}

Disciplines like cognitive science, phenomenology, and psychology identify many different aspects of the Self, but NLP studies: a) have dealt with only a few superficial ones and b) have only employed basic techniques. 
Indeed, while NLP started to employ the correlation between Self-aspects and textual features in various downstream tasks, the Self-aspects employed in, e.g., mental health research and empirical phenomenology, are more varied and nuanced. For this reason, we believe that it would be helpful to identify further and more detailed connections between Self-aspects and textual features, and to develop a model to detect and analyse Self-aspects in text. This could be used by professionals of other disciplines, for instance to analyse patients' reports and transcripts of phenomenological interviews \citep[e.g., see micro-phenomenology;][]{petitmengin2019discovering}. 

To this end, our proposed framework aligns in spirit with existing tools like LIWC \cite{boyd2022development} and VADER \cite{hutto2014vader}. However, unlike these general-purpose approaches, our framework is specifically designed to capture a range of Self-aspects grounded in interdisciplinary theory. Moreover, while LIWC captures psychological correlates at a coarse granularity (e.g., affect, pronouns), we aim to represent structured components of Self-experience.

\section{Research Proposal}
\label{Sec:ResProp}

This Ph.D. proposal seeks to explore the ways of developing a computational model to automatically detect Self-aspects in language. We plan to test the proposed approaches on different case studies from the fields of mental health and empirical phenomenology. Our Research Objectives (ROs) are as follows:

\begin{itemize}
    \item \textbf{RO1)} Detail an ontology of the Self-aspects that would be relevant and sensible for a computational model to detect in text.

    \item \textbf{RO2)} Construct heterogeneous datasets with annotations relative to the identified Self-aspects.

    \item \textbf{RO3)} Define the desiderata of the computational model to detect Self-aspects in text and identify the approaches which would best fulfil them.

    \item \textbf{RO4)} Determine the evaluation approach and the applications for our computational model to detect Self-aspects in text.
\end{itemize}

We plan to produce the following outcomes: a Self ontology with detailing and labelling instructions; heterogeneous annotated datasets; and a set of models to identify Self-aspects in text.

\section{Self Ontology (RO1)}
\label{Sec:SelfTax}

We aim to develop a comprehensive ontology of Self-aspects that are: a) relevant to possible applications, and b) detectable in text data. Each Self-aspect (e.g., Bodily Self) is characterised by different elements (e.g., body ownership and body awareness), each of which is specified in different modes (e.g., body ownership: weak). Some of the Self-aspects investigated are identified through previous studies which developed similar lists or ontologies \citep[e.g.,][]{caporusso2022dissolution, nave2021self}. The ontology, still a work-in-progress \citep[see][]{lukatia}, is built collaboratively by adopting both bottom-up and a top-down approaches. That is to say, we utilise literature detailing the elements and modes of various Self-aspects \citep[e.g.,][]{moore2016sense, serino2013bodily}, along with studies from disciplines like psychology and neuroscience detailing the Self-aspects relevant to the construct of interest \citep[e.g.,][]{petkova2011part}. By way of preliminary illustration (to be refined in later work), consider the various Self-aspects that can be identified in the following excerpts from one of the phenomenological interviews conducted by \citet{caporusso2022dissolution}: ``\textit{I’m very connected with my body.}'' (Bodily Self). ``\textit{The movements are mine, they come from me, there’s nothing separating me from my movements. There isn’t a sense of thinking of having to control all the movements.}'' (Sense of Ownership and Sense of Control). ``\textit{I’m implicitly aware of who I am. (...) Although, it’s not so much about my memories and thoughts, at this moment.}'' (Narrative Self). ``\textit{It’s less about me as me, and more about me as something acting and observing in the moment.}'' (Sense of Agency). ``\textit{I’m having new thoughts, there’s not so much continuity with my past thoughts and my past way of thinking and patterns of thinking.}'' (Thoughts). ``\textit{I’m less caught up in my past Self and I’m more… just something acting in the world.}'' (Relationship with the World). 

Furthermore, we will be meeting with experts from fields that could benefit from applying the final models developed through our framework (e.g., mental health professionals and empirical phenomenologists) to better identify the specific Self-aspects, elements, and modes which could be relevant for their work. While analysing literature and consulting with experts, we will be exploring textual data itself. For each Self-aspect, element, and mode, we will provide a definition, both a positive and a negative example from textual data, and notes to guide the identification and/or distinction among them. Constructing the Self ontology presents various challenges, most of all regarding how the different components relate with each other. For example, most of the aspects and elements, if not all, appear to not be mutually exclusive, and there are aspects (e.g., sense of agency) that could apply to other aspects (e.g., sense of agency over Bodily Self). Moreover, the ontology must navigate differing conceptualisations of the Self across disciplinary traditions. We will address this through an iterative, consensus-driven approach, while remaining anchored in our primary aims of practical applicability and textual detectability.

\section{Datasets (RO2)}
\label{Sec:Data}

The datasets (aiming for at least 10; see Section \ref{Sec:Eva}), which will be annotated with the labels developed (see Section \ref{Sec:SelfTax}), need to vary in type, as it is desired for the model to be able to analyse Self-aspects across different kinds of data. We plan to utilise transcripts from phenomenological interviews, clinical tasks, and structured or unstructured interviews. These will include both existing datasets and newly constructed ones. We aim to utilise datasets from different languages, in order to create a multilingual model. Importantly, all data collection---whether previously conducted or ongoing---is carried out within the scope of pre-approved research projects. Part of the phenomenological interviews data has already been collected (seven subjects), and clinical interviews are being conducted in the context of an existing larger project. 
The annotated datasets will serve as training and testing data, as well as ground truth. The length of the text chunk considered as a labelling instance is determined case by case, based on what is sufficient to meaningfully express the presence of a specific Self-aspect or mode. In general, this can range from a single sentence to a short paragraph, depending on the complexity of the expression.

\subsection{Annotation}

Multiple annotators (e.g., three, possibly the same researchers compiling the Self ontology and the annotation guidelines) will independently annotate the datasets or part of them. The first author, who will take part in and lead the annotation, has experience in conducting qualitative analysis and annotation of textual data, including primarily phenomenological interviews, but also other sources---such as social media posts---with a focus on the Self. In the first phase of the annotation process, the annotators will meet and discuss their decisions, so to come to a similar understanding of the guidelines. This can bring to further adjustments of the guidelines themselves. Inter-annotator agreement will be calculated to assess consistency and reliability of the annotations. Specific annotation training procedures and disagreement resolution protocols will be clearly specified prior to full-scale annotation. A plausible strategy for managing disagreement is majority voting, potentially supported by adjudication from the first author in complex cases. The fact that the annotators may be the same researchers who developed the ontology and guidelines is expected to facilitate consistency and reduce training overhead.
In the case that it proves too expensive to manually label the entire dataset, we will adopt LLMs for automatic annotation of the remaining instances---following an approach similar to that of \citet{caporussophenomenologically}. Specifically, LLMs fine-tuned for instruction following \cite{brown2020language} will be evaluated against a manually annotated subset to ensure quality. Importantly, LLM-based annotations will be used to augment training data for conventional discriminative models, embedding-based retrieval methods, and---in principle---fine-tuning of LLMs, provided such synthetic data is excluded from evaluation (see Section \ref{sec:propapp}). LLMs themselves will be evaluated separately, using only the manually labelled portion of the data to avoid circularity. This ensures a clean separation between training supervision and model evaluation. 

\section{Desiderata (RO3a)}

Here, we discuss our desiderata for the models: interpretability (D1), ground-truth basis (D2), high accuracy (D3), and low computational cost (D4).

Interpretability (D1), which in the context of ML refers to the extent to which a human can understand the internal mechanism of a model leading from input to output \cite{lipton2018mythos, molnar2020interpretable}, 
is to be differentiated from explainability, which often involves post-hoc approximations of a model’s behaviour \cite{molnar2020interpretable}.  
This distinction is particularly crucial for our task for three main reasons. First, the target applications of our framework include implementations in sensitive domains like healthcare. Indeed, in such cases, the use of interpretable ML models is preferable to post-hoc explanations for black-box models, as the latter may be incomplete or misleading and do not ensure transparency, trust, and ethical decision-making \cite{ahmad2018interpretable, amann2020explainability, bohlen2024leveraging, chaddad2023survey, doshi2017towards, ennab2024enhancing, lipton2018mythos, lu2023importance, rudin2019stop, tjoa2020survey}. Some examples of this are studies by \citet{gao2023interpretable} and \citet{wang2023pursuit}. Second, generic explainability approaches are often insufficient in NLP due to the inherent ambiguity, subjectivity, and domain sensitivity of language data, necessitating explanations that align with the linguistic and reasoning norms of specific application areas \cite{mohammadi2025explainability}. 
Some examples of this are studies by \citet{saha2022relational}, \citet{saha2023using}, and \citet{wang2023pursuit}. Third, interpretability is desirable because it enables traceability—the ability to identify the specific passage or linguistic marker that led to a given classification. This is particularly important in applications such as studies based on the analysis of empirical phenomenological interviews, where it is necessary to provide illustrative examples for each identified experiential category \citep[e.g., a specific mode of a Self-aspect; see][]{valenzuela2019analysis}. 

Ground-Truth Basis (D2) requires that model outputs be derived directly from verified, annotated data, rather than inferred through non-transparent or heuristic reasoning \cite{goodfellow2016deep}. Once again, this principle is especially critical in sensitive domains where decisions must be accountable and ethically sound \cite{mittelstadt2019principles, varshney2017safety}, and in NLP, where the inherent ambiguity and subjectivity of language complicate evaluation \cite{hovy2021five}.
In many NLP tasks \citep[e.g.,][]{evkoski2023xai} a degree of approximation is often tolerated in favour of pragmatic utility, and models are evaluated based on what is useful or convincing to downstream consumers. By contrast, in our work, it is strongly desirable that model predictions remain traceable to the actual input provided by us. This grounding is not only central to scientific rigour, but also to ensuring justifiability and trust in use cases such as clinical assessments and the analysis of phenomenological interviews, where outputs may influence human understanding of complex experiences.

Importantly, ground-truth basis is complementary to interpretability. While interpretability focuses on making the model’s decision process understandable, ground-truth basis ensures that its outputs are substantively anchored in verified data rather than emergent patterns from opaque pre-training. Together, these two properties are essential to make computational predictions trustworthy and usable by stakeholders such as clinicians and phenomenologists.

As expected, achieving high classification accuracy (D3) remains a central objective, and considering all the other desiderata, a model with a lower computational cost (D4) is to be preferred. 
Additionally, given the sensitivity of the data, we prioritise tools that guarantee full control over processing and prevent third-party access.

Our main desiderata—interpretability (D1), ground-truth basis (D2), high accuracy (D3), and low computational cost (D4)—form the criteria by which we assess the proposed modelling approaches in Section~\ref{sec:propapp}.

\section{Proposed Approaches (RO3b)}
\label{sec:propapp}

In this subsection, we refer to literature in order to compare the various proposed approaches with regard to each of our desiderata. The proposed approaches are: conventional discriminative models, including traditional AI and neural networks (NNs); generative LLMs, fine-tuned or with few-shot learning; and embedding-based retrieval approaches. 

As the NLP landscape—particularly in relation to LLMs, interpretability, and domain-specific adaptation—continues to evolve rapidly, the methodological choices outlined below are intended as a flexible, revisable framework rather than a rigid pipeline. We anticipate that developments over the course of the Ph.D. will inform and potentially shift our implementation strategies, especially in response to emerging technologies and best practices in ethical, explainable NLP. In line with this adaptable and modular approach, we also propose the investigation of a mixture-of-experts (MoE) architecture.

To train our models, we plan to employ both learned textual features—such as embeddings or TF-IDF representations—and predefined features derived from both previous studies \citep[e.g.,][]{pennebaker2003psychological} and further investigations based on \citet{caporussophenomenologically}'s framework. This hybrid feature strategy supports both data-driven learning and interpretability through grounded linguistic markers.

Preliminary experiments are described in the Appendix \ref{sec:appendix}.

\subsection{Conventional Discriminative Models}

Conventional discriminative models include both traditional ML methods \cite{bishop2006pattern} and NNs \cite{lecun2015deep}. Examples include SVMs \cite{cristianini2000introduction}, logistic regression (LR), decision trees, and feedforward or recurrent NNs \citep[RNNs;][]{goodfellow2016deep} trained for classification purposes. They are often employed in the context of supervised learning, where the model learns from labelled data \cite{murphy2012machine}.




Conventional discriminative models represent a good approach to our goal, assuming the availability of high-quality annotated datasets. Once trained, such models can directly classify a given text instance into predefined categories---such as Bodily Self (BS), Narrative Self (NS), or Self as an Agent (AS)---and further specify the mode for each element (e.g., \textit{bodily ownership: present}; \textit{agency over the body: partial}). Interpretability (D1) in this approach depends largely on the choice of model: while rule-based models like decision trees or LR are inherently transparent, NNs are less interpretable and often require post-hoc explanation methods. Regarding ground-truth alignment (D2), conventional discriminative models are optimal, since their outputs are entirely dependent on the patterns found in the labelled examples. When sufficient and representative training data is available, these models can be very accurate (D3). Furthermore, they can be highly efficient computationally (D4).

\subsection{Generative LLMs}

Generative LLMs \citep[e.g., GPT;][]{radford2018improving} are designed to produce new outputs---in the case of language models, in the form of text---by learning the underlying distribution of the training data \cite{bengio2003neural, radford2018improving}. 

Although flexible, they come with a few challenges. For example, even when a generated response looks plausible, it might be incorrect. This is referred to as \textit{hallucination}, and it is due to the fact that these models generate responses solely based on learned statistical patterns \cite{zhang2022opt}. Additionally, they reflect biases present in their training data and lack transparent mechanisms for interpreting or verifying their outputs \cite{bolukbasi2016man}.

Ideally, generative LLMs will be applied to our task either through prompt-based few-shot learning or via fine-tuning on labelled datasets \cite{wei2022chain, wolf2020transformers}, which generally improves accuracy and control over outputs \cite{howard2018universal}.

While LLMs offer great flexibility and generalisation capabilities, they are not interpretable (D1). Although post-hoc explanation methods like LIME \citep[Local Interpretable Model-agnostic Explanations;][]{alvarez2018robustness, ribeiro2016should} or SHAP \citep[SHapley Additive exPlanations;][]{jin2020bert, lundberg2017unified} can provide some superficial insight, they do not guarantee true transparency or fidelity to the model’s internal reasoning. Furthermore, LLMs are not grounded in ground-truth data (D2). Even when fine-tuned, it remains unclear whether these models' predictions are derived from the data used for fine-tuning or the huge corpora used for pre-training. Furthermore, their outputs can change even from subtle shifts in prompt wording. This affects the consistency and reliability of the model. Accuracy in LLMs is often high \citep[D3; e.g.,][]{wang2025accuracy}, but it depends on prompt design and the complexity of the task. Inconsistent predictions could result from similar inputs, particularly when the classification schema is fine-grained, such as distinguishing between modes of Self-experience. Finally, generative LLMs are computationally expensive (D4).

\subsection{Embedding-Based Retrieval}

Embedding-based retrieval is a type of retrieval-based approach which involves mapping the input into a shared vector space using models such as BERT \cite{devlin2019bert} or Sentence-BERT \cite{reimers2019sentence}. The vector representations of the inputs are compared to the already existing vector space, i.e., the knowledge base  \cite{karpukhin2020dense}. The initial vector space can be fine-tuned to task specific data, enhancing the model performance, and the semantic similarity between the reference and the input texts can be measured via cosine similarity or other distance metrics \cite{cer2018universal, xiong2020approximate}. 



For our purpose, embedding-based retrieval is especially useful in the case that a well-curated repository of annotated examples is available. The model can retrieve similar past instances that have already been labelled, allowing it to infer the classification of the new instance by analogy. While the embedding process itself is not inherently interpretable (D1), the example-based reasoning enabled by retrieval models provides a form of implicit transparency: it is possible to inspect the retrieved examples and their labels to understand the basis of the model's recommendation. This makes the approach more explainable than generative LLMs, although not as transparent as rule-based classifiers. In terms of ground-truth alignment (D2), embedding-based retrieval performs strongly. The model’s decisions are anchored in annotated, verified data, and it does not generate new content but rather identifies the closest match among existing cases. In RAG-style architectures \citep[retrieval-augmented generation;][]{lewis2020retrieval}, this grounding helps reduce---but does not eliminate---the risk of hallucination during generation. Accuracy (D3) depends heavily on the quality and diversity of the dataset: if the database covers a broad range of expressions for different Self-aspects and modes, the model can achieve high classification performance. Computationally, this approach is efficient (D4). Embeddings can be pre-computed, and retrieval operations (e.g., cosine similarity search) are lightweight.

\subsection{Mixture of Experts}

We also plan to explore a mixture-of-experts (MoE) architecture based on the work by \citet{swamy2025intrinsic}, who proposed an interpretable MoE model designed for human-centric applications. In such architectures, different sub-networks---i.e., \textit{experts}, not to be confused with the domain experts mentioned in Section~\ref{Sec:SelfTax}---are selectively activated depending on the input, enabling instance-specific reasoning and the possibility of interpretability (D1) where needed. This design offers a compelling balance between flexibility and transparency: it allows the integration of both interpretable and black-box models within a unified framework. For our purposes, this means we can assign interpretable models to Self-aspect categories where explanation is critical (e.g., clinical applications), while using more complex models for noisier or less constrained categories.

The modular nature of MoE architectures also aligns well with our Self-aspect ontology. Since each expert can be specialised to a distinct subset of Self-aspects or linguistic patterns, this structure supports both conceptual clarity and efficient scalability (D4). Moreover, because only a few experts are activated per instance, the resulting predictions can offer local insight into the decision process, particularly when interpretable experts are selected. Importantly, expert modules trained on annotated data can maintain clear ties to their training supervision, preserving ground-truth basis (D2) at the module level. We believe this architecture is a promising direction to address the trade-off between accuracy (D3) and interpretability across the wide range of Self-related phenomena we aim to model.






\section{Evaluation (RO4)}
\label{Sec:Eva}

\subsection{Intrinsic Evaluation}

To assess the effectiveness of different classification methods for identifying Self-aspects and their elements and modes in text, we will adopt the approach proposed by \citet{demvsar2006statistical} to compare the performance of multiple classifiers across multiple datasets. 
To use this method, a minimum of five different datasets is necessary, although it is recommended to employ at least 10. In the context of this Ph.D., a diverse range of models will be used to perform the classification (see Section \ref{sec:propapp}). Despite their varied architectures and learning paradigms, they all can be evaluated in a comparable way. That is to say, by producing predictions over shared, annotated datasets and assessing them using standard performance metrics such as accuracy, F1-score, or macro-averaged precision and recall.
By using \citet{demvsar2006statistical}'s framework, the evaluation will not only focus on raw performance, but also support robust conclusions about the relative strengths of each approach in the context of supervised Self-aspect classification. This is essential for making informed methodological choices, particularly when weighing the benefits of interpretable and ground-truth-aligned models against those of more flexible and data-driven generative LLMs. 
For the purposes of evaluation, we adopt an instance-based setup, treating each labelled unit (e.g., sentence or utterance) as a classification instance. Future work may explore span-based evaluation to capture finer-grained textual markers of Self-aspect expression. We will also include simple interpretable models and lexicon-based approaches as baselines, to contextualise the performance of more complex systems.

\subsection{Extrinsic Evaluation}

In addition, we plan to evaluate our framework by how useful it proves to be in downstream tasks. As it is likely that different trade-offs of desiderable features are best for different applications, we do not aim to propose one singular model, but a collection of models. They will ideally be implemented in a user-friendly software that will allow the selection of the desired model, along with information and suggestions regarding each of them. Additionally, similarly to LIWC \cite{boyd2022development}, the user will be able to select which Self-aspects to analyse, and to which degree of granularity. It will be possible to determine at which level should the analysis be conducted, e.g., at the sentence, paragraph, or document level.

We intend to conduct at least two case studies in which we will apply one or more of our developed models to different tasks.

In the context of an ongoing project on NLP approaches to cognitive decline, we plan to analyse comparable texts produced by clinical vs non-clinical population by using one or more of our proposed models. In particular, this will serve to test hypothesis on the differences in Self-aspects, but also, potentially, to identify features that could be used to detect cognitive decline.

In the context of the larger attempt to develop a computational framework to support the analysis of phenomenological interviews, one or more of our developed models will be adopted to support the analysis of the phenomenology of the Self, fundamental to most, if not all, experiences. This could help highlight how the Self is experienced differently across an episode \citep[e.g., a dissolution experience;][]{caporusso2022dissolution}, or how it is experienced by different populations, e.g., affected or not by derealisation.

\subsection{Bias Evaluation}

Given the potential impact of our models in sensitive contexts, it is essential to evaluate whether their predictions are affected by social biases. To this end, we plan to adapt and adopt an evaluation strategy inspired by \citet{kiritchenko2018examining}. Specifically, we will test whether the model assigns the same labels to pairs of sentences that are identical in all respects except for a single variation related to a socially salient variable—such as gendered pronouns or racialised names. Any difference in model predictions between such minimal pairs would indicate the presence of bias. Additionally, the presence of bias could be assessed by domain experts during downstream applications.


\section{Conclusion}
\label{Sec:ConclFW}

We presented a proposal to design a computational model capable of detecting Self-aspects in text, grounded in a structured ontology and supported by diverse, annotated datasets curated by us. Our approach bridges conceptual insights from fields such as psychology and phenomenology with empirical techniques in NLP, enabling interpretable and application-oriented analysis of Self in language. Rather than relying on a single architecture, we propose and evaluate a range of computational models---rule-based, embedding-based, and generative LLMs---each assessed in light of desiderata such as interpretability, ground-truth basis, high accuracy, and low computational cost.
By aligning technical development with ethical considerations and application-specific constraints, we aim to contribute not only a functional model, but also a thoughtful framework for the computational study of the Self.

\section{Limitations}
\label{Sec:Lim}

Our work presents various limitations. The Self-aspects specified in our ontology may be insufficient or suboptimal for the range of tasks we intend to address. Additionally, although our datasets are diverse, this may still be insufficient for generalisability---particularly across cultural contexts where expressions of Self may vary significantly. The heterogeneity of the datasets, along with the flexible granularity of labelling units, may also introduce inconsistencies. In terms of implementation, many of the computational approaches we propose require substantial resources, including large volumes of annotated data. The preliminary studies we conducted are limited in scope and therefore insufficient to assess the full feasibility of our framework. Moreover, there is a risk of overfitting to the specific theoretical assumptions embedded in our ontology, particularly if it privileges certain conceptions of the Self over others, potentially narrowing the interpretive scope of our models. Relatedly, the Self is an inherently complex and contested construct, and building an ontology that is both comprehensive and compatible across disciplinary perspectives is itself a theoretical challenge. Reconciling the need for interpretability and ground-truth adherence with high classification performance remains a central challenge in our methodological design. Finally, evaluating our models presents a specific challenge: standard NLP metrics may not fully capture the ability to identify nuanced or context-dependent psychological states. While these metrics enable comparability and rigour, they may only partially reflect the interpretive aims of our framework.

\section{Ethical Considerations}
\label{Sec:Ethi}

As this study relies on existing resources or data collected within the scope of other projects, the ethical considerations for each case are governed by the terms under which the material has been or will be obtained. For corpora accessed through restricted channels, we will comply with all necessary data use agreements and institutional requirements. We are committed to ensuring the anonymisation of all textual inputs prior to model training. Given that both our datasets and the LLMs employed may reflect cultural or demographic biases, we acknowledge the risk of reproducing or amplifying such patterns in model outputs. We emphasise that the computational models developed in this research are intended to function as support tools rather than as standalone decision-makers.

\section*{Acknowledgments}
We acknowledge the financial support from the Slovenian Research Agency for research core funding for the programme Knowledge Technologies (No. P2-0103) and from the project CroDeCo (Cross-Lingual Analysis for Detection of Cognitive Impairment in Less-Resourced Languages; J6-60109). JC is a recipient of the Young Researcher Grant PR-13409. She wishes to thank her supervisors and colleagues---in particular, Matej Martinc, Boshko Koloski, Tine Kolenik, Tia Križan, and Luka Oprešnik. 

\bibliography{custom}

\appendix

\section{Preliminary Experiments}
\label{sec:appendix}

To explore the feasibility of Self-aspect classification in natural language, we conducted a preliminary study focused on the Social Self \citep[SS; ``the self as it is shaped and/or perceived when in an interaction or relationship of sorts with
other people or entities to whom we attribute qualities of an inner life''][]{caporussophenomenologically}, a potential subcomponent of our developing ontology. We selected this category due to its relatively balanced presence in the dataset used and its high inter-annotator agreement during annotation.

\subsection{Dataset and Annotation}

We employed a publicly available dataset of 1,473 diary sub-entries \cite{li2019lemotif}, which we augmented with binary annotations for SS. Annotation combined manual labelling and automated classification using three versions of Gemma2 \cite{team2024gemma}---personalised with psychological and phenomenological expertise. Inter-annotator agreement was assessed via Cohen’s Kappa: 0.80 between human annotators, and 0.84–0.89 between human and model annotators.

\subsection{Experimental Setup}

We trained and evaluated six models using 10-fold cross-validation, combining three different classifiers—support vector machine (SVM), logistic regression (LR), and Naïve Bayes (NB)---with two types of feature representations. The first type comprised learned features, specifically TF-IDF weighted unigrams and bigrams. The second relied on predefined features derived from the LIWC-22 lexicon, specifically those previously identified as correlated with SS \cite{caporussophenomenologically}.
Text preprocessing included converting all text to lowercase, removing punctuation, and applying z-score normalisation to the LIWC-derived features to ensure comparability across feature scales. To interpret the trained models, we employed feature importance techniques tailored to each algorithm: linear SVM coefficients for SVM, SHAP values for LR, and permutation importance for NB.

\subsection{Results}

The best-performing model was the SVM trained on LIWC features, achieving a macro-averaged precision of 0.83 (STD = 0.03), recall of 0.83 (STD = 0.03), and F1-score of 0.83 (STD = 0.03) across 10 folds. These results indicate that it consistently outperformed all other models. Models using learned features (TF-IDF) performed slightly worse overall, with the SVM trained on learned features—the best-performing model among those—achieving a macro-averaged precision of 0.82 (STD = 0.03), recall of 0.81 (STD = 0.03), and F1-score of 0.81 (STD = 0.03). Among the models trained on LIWC features, only NB performed worse than any of those trained on learned features, with a macro-averaged precision of 0.76 (STD = 0.04), recall of 0.75 (STD = 0.04), and F1-score of 0.75 (STD = 0.04). Statistical analysis confirmed the significance of these differences via a Friedman test (statistic = 44.26, p < 0.001) and pairwise Wilcoxon signed-rank tests (adjusted p = 0.03 for several comparisons). Feature importance analyses identified intuitive and interpretable markers of SS, including "we", social referents, affect terms, and pronoun use, aligning with prior findings and theoretical expectations.

\subsection{Implications and Limitations}

This pilot study demonstrates that interpretable models trained on psychologically grounded features can reliably identify expressions of SS in everyday texts. It also confirms the utility of a hybrid human-LLM annotation pipeline, especially in early dataset development.
However, several limitations emerged. Performance is currently limited to binary classification of a single Self-aspect.  
The current study also relies solely on English-language data from a single source, which restricts immediate generalisability.




\end{document}